# Automatic Stroke Lesions Segmentation in Diffusion-Weighted MRI

Using Fluid-Attenuated Inversion-Recovery MRI as Gold Standard

Noranart Vesdapunt

Computer Engineering Department,

Chulalongkorn University, Thailand

*nornart.v@gmail.com*

Assoc. Prof. Nongluk Covavisaruch

Computer Engineering Department,

Chulalongkorn University, Thailand

*nongluk.c@chula.ac.th*

*Abstract*— Diffusion-Weighted Magnetic Resonance Imaging (DWI) is widely used for early cerebral infarct detection caused by ischemic stroke. Manual segmentation is done by a radiologist as a common clinical process, nonetheless, challenges of cerebral infarct segmentation come from low resolution and uncertain boundaries. Many segmentation techniques have been proposed and proved by manual segmentation as gold standard. In order to reduce human error in research operation and clinical process, we adopt a semi-automatic segmentation as gold standard using Fluid-Attenuated Inversion-Recovery (FLAIR) Magnetic Resonance Image (MRI) from the same patient under controlled environment. Extensive testing is performed on popular segmentation algorithms including Otsu method, Fuzzy C-means, Hill-climbing based segmentation, and Growcut. The selected segmentation techniques have been validated by accuracy, sensitivity, and specificity using leave-one-out cross-validation to determine the possibility of each techniques first then maximizes the accuracy from the training set. Our experimental results demonstrate the effectiveness of selected methods.

## I. Introduction

Diffusion-Weighted Magnetic Resonance Imaging (DWI) plays an important role for ischemic stroke detection due to their high sensitivity and fast detection, compared to other modalities [1]. Location and size of ischemic lesions can be extracted by [2] after delicate analysis, which will be useful for decision-making in therapy process and assessment of therapy response. Construction of DWI is done by scanning every voxels to indicate the degree of diffusion of the water molecules in each voxel. The hyper-intense voxels represent the area of ischemic lesions [3], and are inhomogeneous which allow segmentation based on pixel intensity. Natural property of lesions can cause complex shapes, variable sizes, and ambiguous boundaries, therefore, highly experience doctors are required for manual segmentation.

Over the past few years, several methods for stroke lesion segmentation were proposed. Some of those are manual or semi-automatic methods which required excessive labor force and skill. In order to avoid these problems, this project focuses on automatic methods. Past famous automatic methods required ADC map to extract b1000 data [4] or b2000 data [5], however clinical process in the past may not include ADC map into the database. We focus on automatic stroke lesion segmentation which requires only DWI as input.

K-means and Fuzzy C-means (FCM) clustering algorithms are famous segmentation based on clustering techniques [7].

Optimization of k-means clustering algorithm has been made by [8] using hill-climbing algorithm on CIELab histogram resulted as a novel technique called hill-climbing based segmentation. Region growing technique has been tested by [9] and successfully segments the lesions which are suitable for analysis of DWI for classification purpose. Improved version of region growing technique named Growcut was proposed by [10].

Validation of prior methods were done by manual segmentation based on statistical methods, resulted in variation of ground truth caused by operator-dependent. We adopt a semi-automatic segmentation as gold standard aiming to eliminate human error. The selected gold standard was suggested by doctors, based on the similarity of ratio between size of stroke lesion and brain between DWI and FLAIR MRI. We carefully selected gold standards by segmenting stroke lesion areas with Otsu method from FLAIR MRI. We achieve less ambiguous boundaries, and three times higher resolutions than DWI. There are several challenges caused by the difference of modality in the new gold standard such as spatial location error, geometrical distortion, and timeline of lesion evolution

Contributions of this work are as follows:
1. Automatic stroke lesions segmentation algorithms are investigated
2. Four popular techniques for segmentations are identified
3. Gold standards are created by semi-automatic segmentation algorithm based on timeline of lesion evolution
4. Proof of concept is done by leave-one-out cross-validation technique in each algorithm
5. Exhaustive evaluation is performed on six patients in order to find parameters which maximize accuracy

Rest of the paper is as follows, Section II presents the selected methods used for automatic stroke lesions segmentation and theories behind selected gold standard. Section III discusses experiments and results. Discussion on pitfalls in selected gold standard is mentioned in in Section IV. We conclude the paper in section V.

## II. Material and Method

In this research work, three major segmentation techniques including Otsu method, Fuzzy C-means, and Hill-climbing based segmentation were selected to test the accuracy, and created initial labels for Growcut, therefore, six different techniques were tested. Normalization was done before applying segmentation techniques. Registration was done with each DWI to compare with FLAIR and evaluate with accuracy, sensitivity, and specificity.

## A. Candidate Selection

Evolution of stroke lesions in FLAIR and DWI are different due to MR parameters. Experiment by [11] reported some lesions have persistently hyperintense areas on DWI even at 1 month after stroke (see Fig.1), whereas others have become isointense (see Fig.2). If lesions stay persistent, comparison with FLAIR at 1 month is reasonable because [11] stated that signal intensity increases up to day 4 and remains high thereafter on FLAIR (see Fig.3). On day 7, whether lesion stay persistent or not, lesions on both cases are fully developed and comparable to FLAIR.

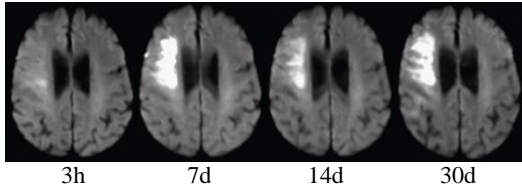

Fig. 1. Evolution of lesion in DWI *(persistent lesion hyperintensity)*

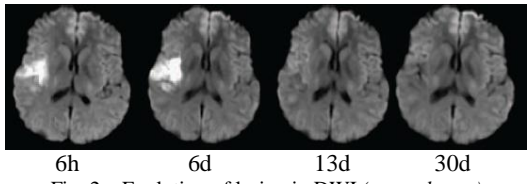

Fig. 2. Evolution of lesion in DWI *(normal case)*

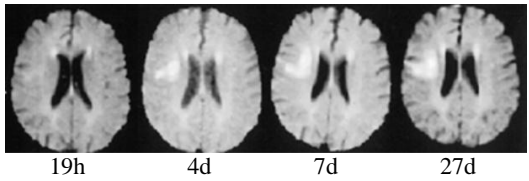

Fig. 3. Evolution of FLAIR

We select 13 DW images from 6 patients which were taken on day 7 or day 30 as training set and 13 FLAIR images from the same patients on the same day as DWI images as gold standard. DW images have been acquired from King Chulalongkorn Memorial Hostpital using Philips Achieva 1.5T. Acquisition parameters used were time echo (TE), 99.5 ms; time repetition (TR), 4272.7 ms; slice thickness, 4.0 mm; gap between each slice, 5.0 mm; total number of slices, 30; b-value, 1000 s/mm$^2$; and pixel resolutions, 224 x 224. The FLAIR images have been acquired from the same machine with time echo (TE), 125.0 ms; time repetition (TR), 11000.0 ms; slice thickness, 6.0 mm; gap between each slice, 7.0 mm; total number of slices, 20; and pixel resolutions, 672 x 672. All images have medical records which have been confirmed by a neurologist and were encoded in 12-bit DICOM format.

Linear transformation, background and noise removal have been applied to DWI and FLAIR MRI before the experiments in order to normalize the images and simplified algorithm computation. Background and noise removal has been done by Philips Achieva 1.5T. We convert images from 12-bit DICOM format to double precision, translating minimum intensity to 0 and scaling maximum intensity to 1.

## B. Otsu Method

Otsu method is one of the most common and successful methods for image thresholding which was introduced by [12]. The objective of method is to find the threshold that minimizes the weighted within class variance. The weighted within class variance ($\sigma_w^2(t)$) is computed as follows:

$$\sigma_w^2(t) = q_1(t_1)\sigma_1^2(t_1) + q_2(t_2)\sigma_2^2(t_2) + ... + q_n(t_n)\sigma_n^2(t_n)$$

n levels of thresholds are produced, $q_n(t)$ is the estimation of class probabilities represented as (1), and Individual class variances ($\sigma_n^2(t)$) are computed by (2):

$$q_n(t) = \sum_{i=t_n}^{t_{n+1}} P(i) \quad (1)$$

$$\sigma_n^2(t) = \sum_{i=t_n}^{t_{n+1}} [i - \mu_n(t)]^2 \frac{P(i)}{q_n(t)} \quad (2)$$

$t$ is the threshold, and $P(i)$ is probabilities of each intensity value, $\mu_n(t)$ is class mean.

We segment DWI by Otsu method into discrete levels from 2 to 20 which resulted in 209 segmented images per DWI and 2,717 segmented images in total.

## C. Fuzzy C-means Clustering

The Fuzzy C-means (FCM) is a well-known clustering method. The objective of method is to find c partition that minimizes the objective function $J_{FCM}$ [13]:

$$J_{FCM} = \sum_{k=1}^{n}\sum_{i=1}^{c} (u_{ik})^q d^2(x_k, s_i)$$

where n is the number of data, $x_k$ is the data set, $c$ is the number of clusters, $u_{ik}$ is the degree of membership of $x_k$ in the $i_{th}$ cluster, $s_i$ is the centroid of cluster $i$, $d^2(x_k,v_i)$ is a distance measure between $x_k$ and $s_i$, $q$ is a weighting exponent.

In this research, we adopt $q$ equals to 2 from [13] and segment DWI into different $c$ between 2-30 which resulted in 464 segmented images per DWI and 6,032 segmented images in total.

## D. Hill-Climbing based Segmentation

Hill-climbing based segmentation is an optimized version of K-means clustering which is simple, fast, and nonparametric. Computation of K-means can be done by a saliency map ($V_k$) as follows [7]:

$$V_k = \frac{1}{|r_k|} \sum_{i,j \in r_k} \sum_s c_{i,j}$$

$r_k$ is the segmented region in pixels, s is the scale of salient regions, and $c_{i,j}$ is the Euclidean distance between the inner and the outer region. Approximation of Euclidean distance is the perceptual differences in CIELab color [7]:

$$c_{i,j} = \| v_1 - v_2 \|$$

$v_1 = [L1; a1; b1]^T$ and $v_2 = [L2; a2; b2]^T$ are the average vectors for the inner and the outer region, respectively. We computed peaks of clusters by climbing each bin until it reaches the highest peak, then select another unclimbed bin to repeat the iterations.

We observe stroke lesions in our training set and infer that stroke lesions always fall on the smallest peak of clusters.

DWI segmentation was done on every cluster between 2-30. In this research, 29 segmented images per DWI were generated, and 377 segmented images in total.

*E. Growcut*

Growcut is a semi-automatic segmentation techniques which required user-labelled background and foreground. Each iteration step allows neighbors to attack the current cell. If they have higher strength, label of the current cell will be replaced by neighbor's label. Decreased function of strength ($g(I_{i,j})$) for each step is defined as [9]:

$$g(I_{i,j}) = 1 - \frac{I_{i,j}}{\max(I)}$$

Where I is pixel value. As the result label and strength of every pixels in the final state are extracted. We construct initialed foreground label by eroding segmentation result from method B-D related to size of stroke lesion :

$$E_n = round(1 + 40 * P_n)$$

$P_n$ is percentage of stroke lesion related to brain area. Erosion type is ball with filter size $E_n$ x $E_n$. Initial background label is created by background pixel of DWI (see Fig.4).

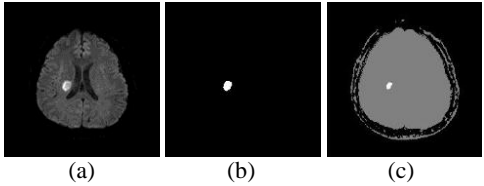

(a)      (b)      (c)

Fig. 4. Initial label created from erosion of stroke lesions segmented by Otsu method. (a) - DWI source, (b) – segmentation by Otsu method, (c) - Initial label, white – 'foreground', black – 'background', gray – 'neutral territory'.

We generated 9,126 initial labels from every segmented images generated by method B-D, and then segment again by Growcut. We observe the relation between strength and segmentation result, and infer that lesion stroke areas always have strength between 0.999-1, thus thresholding techniques was applied for strength 0.999,0.9991,…,0.9999. In total, 91,260 segmented images were generated.

*F. Gold Standard*

Stroke lesions are segmented from 13 FLAIR images by Otsu method using the first proper level to select stroke lesions area. After the segmentation, both FLAIR original and segmented images have been resize using nearest-neighbor interpolation to 224x224. In order to eliminate geometric error in the segmentation area of 2 modalities, image registration technique is applied between DWI and FLAIR MRI to find the transformation matrix. In this research, we use intensity-based image registration with affine transformation. Other type of transformations were tested and failed (see Fig.5).

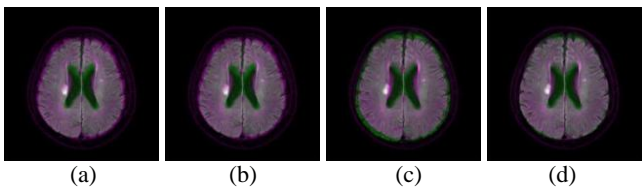

(a)      (b)      (c)      (d)

Fig. 5. Intensity-based image registration between DWI (green) and FLAIR MRI (purple). (a) – translation, (b) – rigid, (c) – similarity, (d) – affine

*G. Performance Measurement*

True positive (TP), true negative (TN), false positive (FP), and false negative (FN) are collected from the assessments. We use popular terms of statistical test including accuracy (3), sensitivity (4), and specificity (5) to measure results.

$$Accuracy = \frac{TP+TN}{TP+TN+FP+FN} \quad (3)$$

$$Sensitivity = \frac{TP}{TP+FN} \quad (4)$$

$$Specificity = \frac{FP}{FP+TN} \quad (5)$$

Comparison of stroke lesions area between segmentation of DWI and gold standard has been done in the region of interest due to overestimated measurement and artifacts.

### III. EXPERIMENTS AND RESULTS

*A. Leave-one-out Cross-Validation*

Due to the lack of test cases, proof of concept has been done for each method in section II to verify the methods. 12 DW and FLAIR images were used to find parameters that maximizes accuracy with sensitivity and specificity higher than 90 percent. Results are shown in TABLE I.

TABLE I. RESULT OF LEAVE-ONE-OUT CROSS-VALIDATION

| method | *Accuracy* | *Sensitivity* | *Specificity* |
|---|---|---|---|
| Otsu method | 91.72± 6.08 | 91.54±7.24 | 91.99±8.00 |
| Fuzzy C-means | 91.29±2.44 | 89.46±11.1 | 91.56±2.96 |
| Hill-climbing | - | - | - |
| Otsu method & Growcut | 92.09±5.54 | 93.5±6.85 | 92.48±6.56 |
| Fuzzy C-means & Growcut | 96.01±3.07 | 89.95±17.2 | 96.70±2.58 |
| Hill-climbing & Growcut | 89.51±13.9 | 93.78±6.30 | 88.58±17.2 |

Segmentation by Hill-climbing based segmentation does not satisfy minimum requirement of sensitivity and specificity, while other methods produce promising results.

*B. Finding Parameter Value*

After proof of concept is done, we setup an experiment to find the parameters value which maximizes the accuracy and satisfies minimum requirement of sensitivity and specificity. Results are shown in TABLE II.

TABLE II. RESULT OF FINDING PARAMETER VALUE

| method | *Parameters* | *Accur-acy* | *Sensi-tivity* | *Speci-ficity* |
|---|---|---|---|---|
| Otsu method | 17,10 | 91.721 | 91.543 | 91.987 |
| Fuzzy C-means | 29,28 | 90.973 | 92.560 | 90.538 |
| Otsu method & Growcut | 16,14/0.9991 | 92.966 | 96.112 | 92.278 |
| Fuzzy C-means & Growcut | 28,28/0.9999 | 95.872 | 90.576 | 96.422 |
| Hill-climbing & Growcut | 22/0.9995 | 95.500 | 91.600 | 96.204 |

Parameters are identified as follows: Otsu method – number of levels and threshold level, Fuzzy C-means clustering – number of clusters and selected cluster, Hill-climbing based segmentation – number of bins, Growcut – threshold of strength.

Highest accuracy with sensitivity and specificity higher than 90 percent is received from Fuzzy C-means clustering with Growcut. Other methods except Hill-climbing based segmentation produce high accuracy, sensitivity, and specificity.

*C. Speed*

Processing time of Fuzzy C-means clustering with Growcut (number of clustering – 28, selected cluster – 28, strength threshold – 0.9999) testing on DWI (resolution - 224x224, 12 bit, DICOM format) on Intel ® Core™ i3-3110M with frequency 2.40 GHz is 94.772 second. 90.041 second is consumed by Fuzzy C-means clustering, and 1.182 second is used by Growcut. Time consumption in this method is too large for real time application.

On the other hand, only 0.274 second is needed in Otsu method (number of levels – 17, threshold level – 10) testing on the same image and machine. It's reasonable to use Otsu method as computer-aided diagnosis (CAD) due to the difference of accuracy between 2 methods is only 4.151 percent while 99.71 percent of time consumption is reduced.

## IV. DISCUSSION

Our experimental results reflect high accuracy of selected methods which are tested on FLAIR MRI as gold standard, however, verifications are needed to analyze the effectiveness of FLAIR MRI as gold standard. We realize 2 important researches which are needed to investigate further.

*A. Evolution of stroke lesions in DWI and FLAIR*

Similarity of ratio between size of stroke lesion and brain between DWI and FLAIR MRI was analyzed by observation on small test cases. Extensive testing is required to analyze the error between size of stroke lesions in DWI and FLAIR MRI after image registration.

Timeline of stroke lesions evolution reported by [11] assigned the first date of symptom onset. These dates are provided by diagnoses which are required highly experience doctors to estimate accurate dates from the patients. However, these errors will slightly affect segmentation results because only small changes could be happened around day 7 or day 30 of symptom.

*B. Stroke lesions segmentation in FLAIR MRI*

We carefully segment stroke lesion areas from FLAIR MRI by Otsu method using the first proper level. Due to the properties of FLAIR MRI, higher resolution and unambiguous boundaries make it easier to segment stroke lesions areas. We break down the problem of stroke lesions segmentation in DWI, which is hard to acquire, into stroke lesions segmentation in FLAIR MRI which has higher image quality. Instead of using Otsu method, an extensive testing on segmentation techniques should be done to identify the most proper technique for segmentation.

If we achieve such technique, further training for stroke lesions segmentation in DWI can be done without highly experience doctors to carefully segment and verify the gold standards.

## V. CONCLUSION

This paper presents automatic stroke lesions segmentation in DWI. Comparison of four popular methods is done including Otsu method, Fuzzy C-means clustering, Hill-climbing based image segmentation, and Growcut. Pre-processing is performed by linear transformation, background and noise removal. Evaluation is achieved by testing on FLAIR MRI as gold standard. Proof of concept is carried out for segmentation methods validation, which resulted in high accuracy, sensitivity, and specificity except Hill-climbing based image segmentation. Another experiment identifies Fuzzy C-means clustering with Growcut as the method with highest accuracy but lack of speed for real-time application. We conclude that Otsu method is a suitable method for CAD, and Fuzzy C-means clustering with Growcut is the best method to maximize the accuracy.

ACKNOWLEDGMENT

This research was collaborated by Chulalongkorn Memorial Hostpital. We thanks Assoc. Prof. Dr. Sukalaya Lerdlum, and Dr. Yuttachai Likitjareon for biomedical advices.

REFERENCES

[1] P. Barber et al. "Identification of major ischemic change: diffusion-weighted imaging versus computed tomography," Stroke, vol. 30(10), pp.2059–2065, 1999.

[2] Lovblad K.O., Laubach H.J., Baird A.E., et al. "Clinical experience with diffusion-weighted MR in patients with acute stroke," AJNR Am J Neuroradiol, vol. 19(6), pp. 1061-6, 1998.

[3] P.W. Schaefer, P.E. Grant, R.G. Gonzalez, "State of the art: diffusion-weighted mr imaging of the brain," Annual meetings of the radiological society of north america (RSNA), 2000.

[4] K. Bhanu Prakash et al. "Automatic processing of diffusion-weighted ischemic stroke images based on divergence measures," CARS, vol. 3(6), pp. 559–570, 2008.

[5] Shashank Mujumdar, R. Varma, L.T. Kishore, "A novel framework for segmentation of stroke lesions in diffusion weighted mri using multiple b-value data," ICPR (IEEE), pp 3762-3765, 2012

[6] Rubens Cardenes, Rodrigo de Luis-Garcia, Meritxell Bach-Cuadra, "A multidimensional segmentation evaluation for medical image data," J. computer methods and programs in medicine, vol. 96, pp. 108-124, 2009.

[7] Shailesh Kochra, Sanjay Joshi, "Study on hill-climbing algorithm for image segmentation," International Journal of Engineering Research and Applications, vol. 2, pp. 2171-2174, 2012.

[8] N. Mohd Saad, S.A.R. Abu-Bakar, Sobri Muda, M. Mokji, A.R. Abdullah, "Fully automated region growing segmentation of brain lesion in diffusion-weighted MRI", IAENG international journal of computer science, vol. 39:2, IJCS_39_2_03, 2012.

[9] Vladimir Vezhnevets, Vadim Konouchine, "Growcut - interactive multi-label n-d image segmentation by cellular automata," Accepted to Graphicon-2005.

[10] Carly S. Rivers, et al. "Persistent infarct hyperintensity on diffusion-weighted imaging late after stroke indicates heterogeneous, delayed, infarct evolution", Journal of the american heart association, vol. 37, pp. 1418-1423, 2006.

[11] Maarten G. Lansberg, et al. "Evolution of apparent diffusion coefficient, signal intensity of acute stroke", AJNR Am J Neuroradiol, vol. 22, pp. 637–644, 2001.

[12] N. Otsu, "A threshold selection method from gray-level histogram", IEEE Trans. systems man, and cybernetics, vol. 9, pp. 62-66, 1979.

[13] Yong Yang, Shuying Huang, "Image segmentation by fuzzy c-means clustering algorithm with a novel penalty term", Computing and informatics, vol. 26, pp. 17-31, 2007